\documentclass[letterpaper]{article}

\usepackage{natbib,alifeconf}  
\usepackage{url,hyperref,cleveref}
\usepackage{booktabs}
\usepackage{microtype}
\usepackage{caption}
\usepackage{enumitem}
\title{Large Language Models as In-context AI Generators for Quality-Diversity}

\author{
    Bryan Lim$^{1}$,
    Manon Flageat$^{1}$, \and
    Antoine Cully$^1$ \\
    \mbox{}\\
    $^1$Imperial College London, United Kingdom \\
    bryan.lim16@imperial.ac.uk
} 

\begin{document}

\maketitle

\newcommand{\algoname}{In-context QD}

\begin{abstract}
    Quality-Diversity (QD) approaches are a promising direction to develop open-ended processes as they can discover archives of high-quality solutions across diverse niches. 
    While already successful in many applications, QD approaches usually rely on combining only one or two solutions to generate new candidate solutions. 
    As observed in open-ended processes such as technological evolution, wisely combining large diversity of these solutions could lead to more innovative solutions and potentially boost the productivity of QD search.
    In this work, we propose to exploit the pattern-matching capabilities of generative models to enable such efficient solution combinations. 
    We introduce In-context QD, a framework of techniques that aim to elicit the in-context capabilities of pre-trained Large Language Models (LLMs) to generate interesting solutions using few-shot and many-shot prompting with quality-diverse examples from the QD archive as context.
    Applied to a series of common QD domains, In-context QD displays promising results compared to both QD baselines and similar strategies developed for single-objective optimization.
    Additionally, this result holds across multiple values of parameter sizes and archive population sizes, as well as across domains with distinct characteristics from BBO functions to policy search. 
    Finally, we perform an extensive ablation that highlights the key prompt design considerations that encourage the generation of promising solutions for QD.
\end{abstract}

\section{Introduction}

Open-ended processes are systems that continuously generate novel and interesting outcomes and products.
They have tremendous potential as tools for creativity or as invention generators, and could drive the next generation of scientific discovery and lead to major AI innovations~\citep{stanley2019whyopen}.
However, designing algorithms and systems which are truly open-ended and capable of never-ending creativity and complexity still remains a challenge~\citep{stanley2017oegrandchallenge}.
One of the challenges in open-ended search is generating solutions that are novel and better, given knowledge and solutions that already exist. 

We take inspiration from how humans invent and innovate and how the creative process of invention often involves combining many diverse concepts together.
Humans rely on a diversity of past solutions in the form of other inventions, artifacts and knowledge-bases such as books and code-bases to collectively and creatively generate new inventions. 
For example, the development and invention of a device like the smartphone was built upon a large diversity of prior generations of inventions and engineering from screens (material science), power supplies (chemistry), processing chips and many more.
Provided these building blocks, the ingenuity and creativity of humans to effectively ingest and extract the key elements from this diversity of examples is also critical to making such breakthroughs and discoveries. 
More often than not, many of the modern technologies follow this same observation and are rarely developed in isolation and from scratch.
While this recombination is characteristic of processes such as cross-over in natural evolution, these usually involve combining only two parent solutions. 
Effectively combining and using a large number of past inventions is not trivial to achieve. 
Our work looks at replicating this open-ended process of invention and innovation observed in cultural and technical evolution by (i) using foundation models to effectively ingest a large diversity of solutions to generate solutions that are both better and more novel, and (ii) using Quality-Diversity to maintain and provide these models with many diverse and high-quality examples as context for generation.

Quality-Diversity (QD)~\citep{pugh2016quality, cully2017quality, chatzilygeroudis2021quality} algorithms are an effective way to collect and maintain a diversity of the best solutions seen and has been a step towards open-ended algorithms. 
Instead of just searching for a single optimal solution in conventional optimization, QD searches for a diversity of high-performing solutions.
A QD approach is useful when one wants a diversity of good options and solutions at the end of search, but can also be implicitly effective in the conventional optimization setting by providing stepping stones to a more optimal solution~\citep{gaier2019quality, clune2019ai}, that otherwise might have been discarded.
This has benefits across many applications such as robotics~\citep{cully2015robots}, procedural content generation~\citep{gravina2019procedural}, engineering design~\citep{gaier2018data} and many more.



In QD, new solutions are commonly generated either through random mutations and variations, or optimization algorithms such as evolution strategies (ES) or gradient descent (i.e. policy gradients). 
These generation methods use the existing archive of elites mainly as starting points, either as parents for mutations, or for initialization of optimization procedures such as gradient descent. 
Existing QD approaches rarely consider using the entire archive (or even multiple solutions from the archive) to generate new solutions, while having this additional information could have great benefits and be more effective for the generation process.

We propose to address this challenge by using large transformer models such as Large Language Models (LLMs).
This is possible due to the in-context learning capabilities of large transformer models~\citep{brown2020language}, where given prompts of arbitrary examples and demonstrations of a task, these models can extract patterns effectively to generate completions and even improvements~\citep{brown2020language, laskin2022context, generalpatternmachines2023}.
This work introduces in-context AI generators for QD (In-context QD), where new solutions are generated using an LLM by passing a representative subset of the archive of elites into the context of the model. 
Instead of just starting points, we use the in-context capabilities of pre-trained LLMs to ingest a larger diversity of high-performing solutions from the archive to improve the generation of solutions by leveraging both the pattern-matching and generative capabilities of LLMs. Specifically, we introduce few-shot prompting strategies for QD by providing solutions from the archive of elites as examples in the context when prompting the LLM. 
QD lends itself nicely to in-context learning by providing a diverse set of high-quality and diverse examples for the model to potentially generate more interesting ideas and solutions. 


We show that by careful construction of template, context and queries of the prompt, \algoname{} can effectively generate novel and high-quality solutions for QD search over a range of parameter search space dimensions, and archive sizes.
We rigorously demonstrate this over a range of QD benchmarks with ranging characteristics such as benchmark BBOB functions, redundant robotic arm control and parameterized policy search for a robot.
In most cases, \algoname{} performs better than MAP-Elites across metrics especially in finding regions of high-fitness.
Additionally, we conduct a thorough study over components of \algoname{} and highlight the importance of the choice of context template, context structure and context size.

\begin{figure*}
    \centering
    \includegraphics[width=0.87\linewidth]{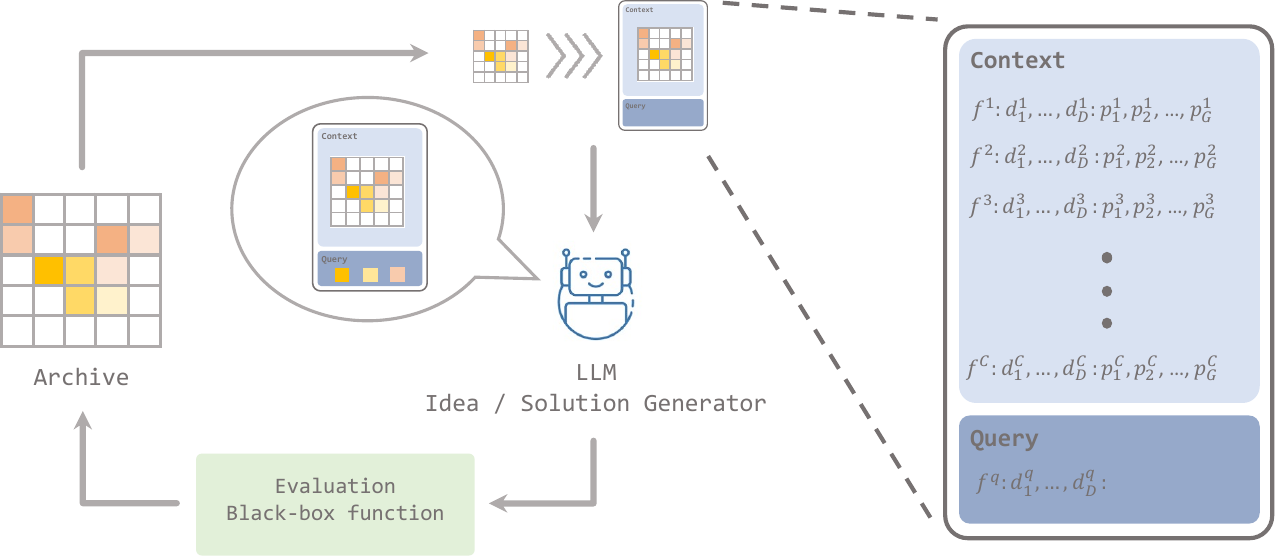}
    \caption{Overview of \algoname{} using LLMs as idea and solution generators. The LLM observes the current state of the archive which contains high-quality and diverse solution examples and uses this to generate new solutions to be evaluated and considered for addition to the archive. The prompt consists of \textit{context} and \textit{query} portions which are concatenated together.}
    \vspace{-4pt} 
    \label{fig:in-context-qd}
    \vspace{-2mm}
\end{figure*}

\section{Background Related Work}

\subsection{Quality-Diversity and MAP-Elites}
Drawing from natural evolution, Quality-Diversity (QD) optimization actively seeks a variety of solutions to a given problem, akin to species with diverse characteristics achieving the same objective in the natural world. 
In other words, instead of looking for one objective-maximizing solution, QD searches for a collection of diverse and high-performing solutions. 
To do so, most QD approaches rely on a fitness $f$ that quantify the quality of solutions and a set of pre-defined characteristics that allow quantifying their diversity, referred to as features $\textbf{d}$.

Among existing QD approaches, the most commonly used one is MAP-Elites~\citep{mouret2015illuminating}, which is the focus of this work. 
MAP-Elites maintains an archive of solutions that obtains its structure by splitting the feature space into a multidimensional archive of $C$ cells. Each cell of the archive defines a feature-space neighborhood (i.e. niche) where one solution can be kept. 
The solutions kept in the cells are referred to as elites and the set of all elites is returned as the final collection of solutions of the algorithm. 
The use of an archive guarantees a good coverage of the feature space and thus a large diversity of the final elites according to the feature dimensions. 
After initialization with random solutions, a MAP-Elites loop follows the standard cycle of selection, mutation, evaluation and addition to the archive. 
The selection of parents is done by uniformly sampling from the existing elites of the archive, to encourage quality and diversity. 
The addition to the archive follows two rules: (1) if the new solution fills in an empty cell, it is always added to that cell, and (2) if there is already an elite in the cell, the new solution only replaces it if it has a higher fitness $f$.
These conditions ensure that added solutions either improve the diversity or quality of the archive.

\textbf{Variations and Optimizers.}
In MAP-Elites, the mutation can be done using any common mutation operator such as standard GA~\citep{mouret2015illuminating} or specific operators tailored for QD~\citep{vassiliades2018iso}.
Previous work has also proposed to use gradient-informed mutation when available~\citep{fontaine2021differentiable, nilsson2021policy, pierrot2022diversity}, or mutation based on evolution strategies~\citep{colas2020scaling, fontaine2020covariance, flageat2023multiple}.
In comparison, our work leverages the pattern completion abilities of LLM to generate the next solution given the large diversity of the archive of elites.
While LLMs have also been used as mutation operators for their ability to tackle text-based and program-based QD domains\citep{lehman2023evolution, bradley2023quality}, we focus and demonstrate these capabilities on general QD BBO optimization problems, other than text or code and develop prompt construction strategies specifically for QD.
Also close to our work, BOP-Elites~\citep{kent2020bop} and DDE-Elites~\citep{gaier2020discovering} both learn a representation of the current archive content and use this representation as a mutation operator to generate new relevant offspring. 
They rely on a Gaussian Process and a Variational Autoencoder model respectively. 
While our work also uses a representation of the archive content to generate relevant offspring, it relies on the in-context capabilities of pre-trained foundation models and does not require training a model online during the optimization. 
We show in the following that the LLM manages to generate relevant offspring without requiring any online specific training.


\textbf{Model-based QD.}
Model-based QD are a family of QD methods that commonly use learnt models as surrogate models to increase the sample efficiency of search~\citep{gaier2018data, keller2020model, lim2022dynamics, zhang2023using} or in domains in which model outputs are good proxies~\citep{bradley2023quality, ding2023quality}.
LLMs are also models, but instead of using these models for their predictive capabilities, our approach uses them for their generative capabilities.

\subsection{Language Models as Optimizers}
Inspired by evolutionary computation, LLMs have been very recently used as mutation or variation operators, especially in text-based domains, ranging from modifying sentence sentiment to generating mathematical expressions, code and programs~\citep{lehman2023evolution, meyerson2023language}. 
In fact, when searching and optimizing in the space of text and prompts, using evolution-based methods is one of the more popular and effective approaches~\citep{yang2023large, fernando2023promptbreeder} as they are black-box optimization algorithms.
Additionally, LLMs can act as optimizers and work over sequences of integers~\citep{generalpatternmachines2023, yang2023large}, as well as problems that can be represented as code or programs~\citep{chen2024evoprompting, romera2024mathematical, trinh2024solving}.

Prior work using LLMs as optimizers has mainly focused on either (i) conventional optimization problems where the goal is to find a single solution instead of a QD problem~\citep{yang2023large, generalpatternmachines2023, meyerson2023language, zhang2023using}, or (ii) focuses on the QD problem but uses the LLM in a way that does not explicitly account for both the fitness and feature in the prompt and has mainly been used in subjective text-based domains~\citep{bradley2023quality, samvelyan2024rainbow}.
There is preliminary work which uses LLMs for more generic evolutionary optimization~\citep{liu2023large} but considers the case of conventional optimization and requires very detailed task information and descriptions.
Closest to our approach is concurrently released work which also studies prompting strategies but targeting recombination operators for Evolution Strategies (ES)~\citep{lange2024large}, where the LLM generates the next mean statistic of an ES which solutions are sampled from.
Instead, our work considers the general QD problem which contains the added dimensions of diversity.
Additionally, our approach taps into the curated solutions in the archive of elites to build the context and studies in-context prompt construction approaches for the QD problem and directly generates sample solutions.

\section{Method}

In QD, the goal is to find a diversity of high-quality solutions. 
This provides two dimensions of optimization, fitness and novelty/diversity, compared to conventional single-objective optimization approaches which only consider fitness.
A common approach when using LLMs as optimizers is to sort the sequence in an increasing order of fitness and provide a query of higher value, allowing the model to complete the sequence~\citep{yang2023large, generalpatternmachines2023}.
However, in QD, we are not concerned with just finding a solution that has a high fitness, we are also interested in covering the feature space with high-scoring solutions.
Figure~\ref{fig:in-context-qd} shows an overview of our approach to this, \algoname{}.
The LLM takes as input the current state of the archive, and uses this as a prompt to generate new solutions.
The generated solutions are then considered for addition to grow the quality and diversity of the archive as in MAP-Elites.
In the following, we explain how we represent the solutions and build a prompt that consists of the \textit{context} and the \textit{query}.
This provides a general and modular framework for which to use LLMs as AI generators for Quality-Diversity.

\subsection{Prompt Template}

To include both fitness and all the dimensions of features, we use a prompt template that takes the form: 
\begin{equation}
    f^i: d^i_1, ..., d^i_G: p^i_1, ..., p^i_D
\end{equation}

where i represents the index of a solution, $G$ represent the total dimensions of the features and $D$ represent the dimensions of the parameter space.
The fitness $f$, features $\textbf{d}$ and parameters $\textbf{p}$ are separated by a colon ":" while each value is separated by a comma ",".
Each solution $i$ is then separated by a line break "$\backslash n$".
This representation allows us to consider all elements of the QD problem.
We follow prior work~\citep{generalpatternmachines2023, lange2024large} and represent the features $\textbf{d}$ and parameters $\textbf{p}$ as integers. 
This is due to how numbers/digits are tokenized in LLMs, where representing a large combination of digits stacked together as a float/decimal might be tokenized, broken up and ingested in a non-uniform and confusing manner for the model~\citep{beren2024integer, singh2024tokenization}.

\subsection{Building Context}
The solutions maintained in a QD archive lend themselves nicely to few-shot and many-shot prompting of LLMs as the diversity and quality of solutions provide a good set of examples and context to refer to in order to generate better ideas and solutions.
Ideally, the entire archive could be used as the context.
However, given the context size limit of many open-source LLMs, we use a sample/subset of solutions of the archive as an approximation of the archive to construct the context.
If provided with models with larger context or problems with lower archive sizes, the entire archive could be used as the context.
The context size is a hyper-parameter and can take values up to the archive size $C$.
Yet, despite this temporary limitation, we can already demonstrate the effectiveness of this proposed approach.

In building the context, there are two considerations: the \textbf{\textit{content}} and the \textbf{\textit{structure}} of the context.
As explained above, the \textbf{\textit{content}} of the context is a subset representation of the archive. 
We uniformly sample solutions from the archive to form the content of the context.
An increasingly larger context of solutions sampled would give us an increasingly better representation of the archive culminating in the entire archive itself.
\citet{meyerson2023language} suggest that few-shot prompting can be viewed as Estimation of Distribution Algorithms (EDAs).
In our QD setting, by providing more solutions from the archive in the context using our prompt template, we provide a good approximation of the patterns and the distributions of the fitness-feature space.
This provides a better context for the language model especially when queried for desired features.

Next, we look at the \textbf{\textit{structure}} of the context. 
As observed in prior work~\citep{generalpatternmachines2023, yang2023large} and also demonstrated in our work, the structure of the context also matters to provide a clearer heuristic and pattern to the model.
There are many possible strategies for this.
We introduce sorting according to distance from the query features (see next section) where solutions in the context are ranked from the furthest to the closest to provide a heuristic in the direction of the features.
Alternatively, we can also opt for the tried and tested strategy of sorting the context based on the fitness scores.
However, this might not always be effective in all QD domains as the search for diverse solutions is a big component of QD.
Other options could include sorting the solutions randomly, or according to the certain dimensions of the features, or to other scores of interest.
Through the query strategy, we will see why this makes sense as an overall approach to QD.

\subsection{Query Strategy}

\begin{table}[t!]
\footnotesize
    \centering
    \begin{tabular}{l|c}
    \toprule
    Configuration & Options \\
    \midrule
    \addlinespace
    Prompt Template  &  (LMX, Fitness, Feature, QD) \\
    Context Size  &  [0, $C$] \\
    Context Structure  &  (Sort Fit., Sort Cell Dist., Random)    
    \\
    Fit. Query Strategy & Context Improvement  \\
    Desc. Query Strategy & (Empty, Uniform)  \\
    \bottomrule
    \end{tabular}
    \caption{Variants of context building and query strategies.}
    \label{tab:llmqd-variants}
    \vspace{-2mm}
\end{table}

To get the language model to generate a solution that will improve the quality and diversity of our current archive, we take the "just ask"~\citep{jang2021justask} approach.
We will "ask" the LLM for quality and diversity, or in particular, solutions that are novel w.r.t to the archive and are high-quality.
To do this, we append a query fitness $f^{q}$ and features $d^{q}_1, ..., d^{q}_G$ to the context prompt (see Figure~\ref{fig:in-context-qd}) to elicit a pattern or sequence completion behavior from the model, based on the context.

\begin{figure*}[t]
    \centering    \includegraphics[width=0.8\linewidth]{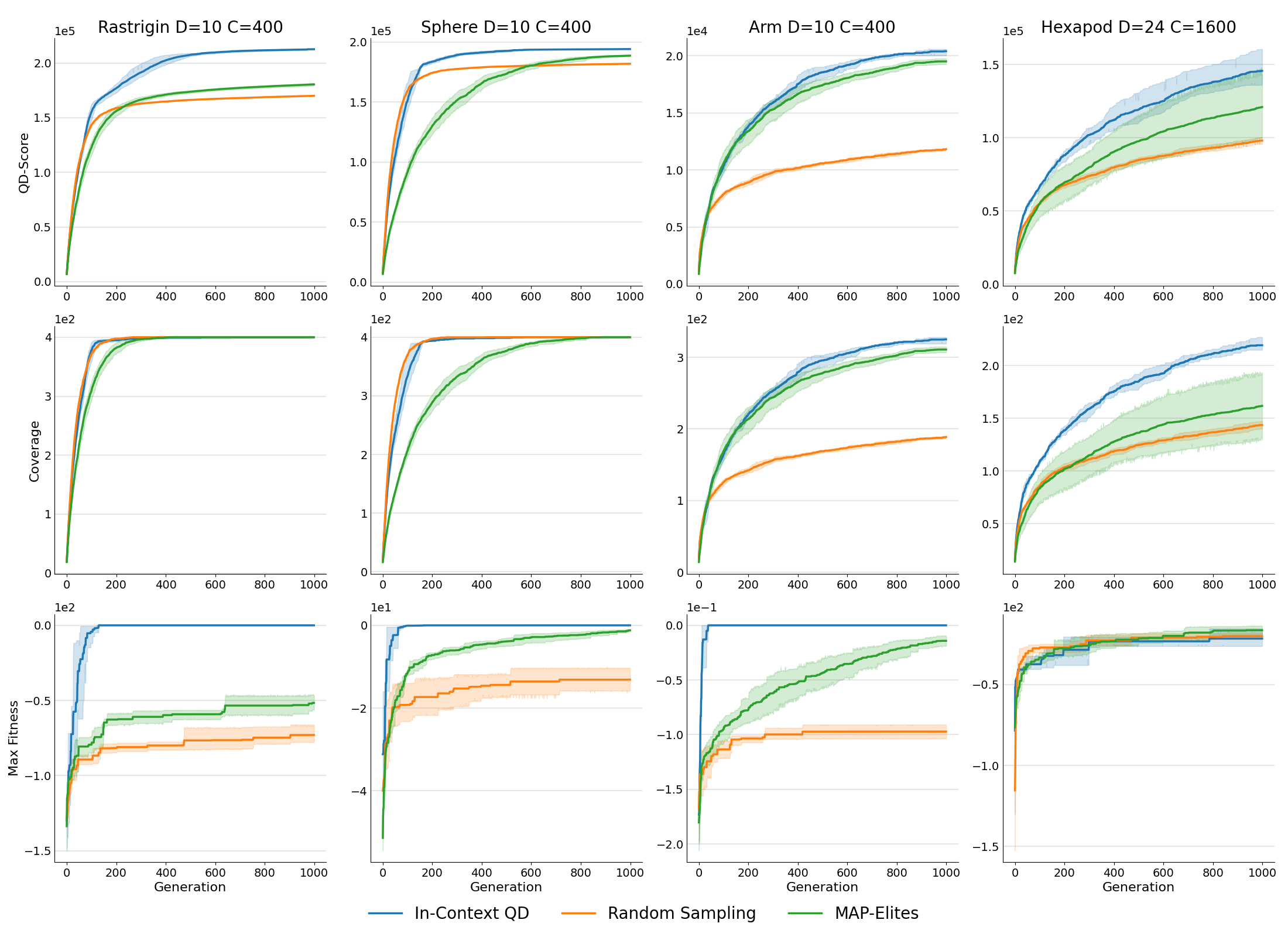}
    \captionsetup{skip=1pt}
    \caption{Performance comparison over QD-Score, Coverage and Max-Fitness across all tasks considered where $D$ is the parameter space dimensions and $C$ is the number of niches in the archive. The results are averaged over 5 independent runs.}
    \label{fig:in-context-qd-results}
    \vspace{-3mm}
\end{figure*}

We pick a fitness query $f_q$ approach that complements and fits the context, which is a sampled representation of the QD archive in the style of the prompt template. 
We set a fitness query $f_q$ that is incrementally larger ($\sim20\%$) than the best fitness solution in the context.
This fitness query strategy is referred to as context improvement.
Such a query always ensures a reasonable increase in fitness performance given the context sequence to ensure a systematic trend and pattern in the prompt.

The selection of features used in the query $d^q$ cannot be treated entirely the same as with the fitness.
In the case of features, the goal is to cover the entire feature space and not find an optimal point in the space by moving in a specific direction of a feature.
While expanding the horizon and boundary of the archive is needed, it is also just as important to fill in "holes" in the archive to maximize the coverage of the feature space.
An unbiased approach to this is to uniformly sample the centroids of the cells in the archive to obtain query features $d^{q}_1, ..., d^{q}_G$. 
We refer to this approach as "Uniform".
We also propose a strategy which queries features from cells that are empty first, after which we default to sampling features uniformly from the cells in the archive when the number of empty cells becomes less than the batch size. 
The centroids of the empty cells and sampled cells are used for the queries. We refer to this strategy as "Empty".

Overall, this proposed context-building approach is modular, composable and easily amendable to new templates, context-building or query strategies.
Table~\ref{tab:llmqd-variants} summarizes the different options available when applying \algoname{}.

\section{Experimental Setup}

We evaluate our approach on a range of benchmark QD optimization tasks.
As \algoname{} relies heavily on pattern and sequence matching and completion, we highlight characteristics of how the parameters are related to the fitness and feature in each task.
We first evaluate on two popular functions from the  Black-Box Optimization Benchmark (BBOB)~\citep{hansen-bbob:inria-00462481}: the \textbf{Rastrigin} (Rosenbrock function) and \textbf{Sphere} tasks as done commonly in QD literature~\citep{mouret2015illuminating, fontaine2020covariance}.
The features in these synthetic QD tasks are directly tied to the parameter space making exploration of the feature space relatively simple, where random uniform sampling over the parameter space can result in high coverage.
Next, we consider the redundant planar robotic \textbf{Arm} task~\citep{mouret2015illuminating, cully2015robots} where the parameters represent the angles of each joint in the system.
Here, the features are the final end-effector position, which is less trivial and cannot be directly inferred from the parameters.
For these three tasks, we evaluate our approach across different parameters space dimensions $D \in \{5, 10\}$.
Finally, we also consider a policy search task where search is over $D = 24$ parameters, of a locomotion controller for a \textbf{Hexapod} robot~\citep{cully2013behavioral, chatzilygeroudis2018reset}.

We compare in-context QD against a Random Sampling baseline which samples the parameter space uniformly and MAP-Elites~\citep{mouret2015illuminating} with the iso-line variation operator~\citep{vassiliades2018iso}.
We evaluate the (i) maximum fitness score found throughout search, (ii) the coverage, measured by the number of filled cells in the archive and (iii) the QD score which is the sum of fitness of all the solutions in the archive.
For all our algorithms and baselines, we use a batch size of $10$ at each generation, and run for 1000 generations.
We also investigate the performance across different feature space sizes $C \in \{400, 1600\}$ by changing the resolution of the MAP-Elites archive $C \in \{20^2, 40^2\}$ to ensure the generality of our approach.
To be able to run in-context QD on just a single GPU device, we use Mistral-7B-v0.1~\citep{jiang2023mistral}, an open-source 7B parameter language model.

\section{Results}

\begin{figure}[t]
    \centering\includegraphics[width=\linewidth]{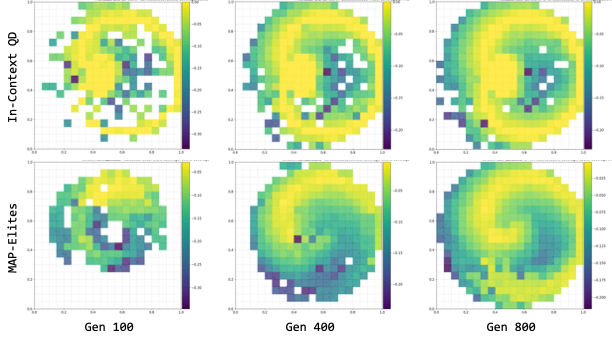}
    \captionsetup{skip=2pt}
    \caption{Evolution of archives across generations showing the different optimization paths and dynamics of \algoname{} which can quickly exploit patterns for regions of high-performance.}
    \vspace{-4mm}
    \label{fig:archive-optimization-dynamics}
\end{figure}

Figure~\ref{fig:in-context-qd-results} shows that \algoname{} is capable of performing QD optimization and outperforms both Random Sampling and MAP-Elites baselines on the synthetic BBOB tasks. 
This is interesting as both Random Sampling and MAP-Elites have shown to be very effective in low-dimensional optimization problems despite their simplicity, especially on these synthetics tasks~\citep{mouret2015illuminating}.
We can observe the effectiveness of \algoname{} in finding the global optima in these QD problems, quickly extracting the patterns that lead to high fitness scores.
This result suggests that \algoname{} maintain the ability to find the best solution in the conventional optimization settings as observed in other work~\citep{yang2023large, generalpatternmachines2023, lange2024large}.
More interestingly, our QD prompting strategy allows the LLM to also effectively extract patterns in the feature distribution to generate more solutions that are diverse and cover the feature space more efficiently.
In these BBOB functions, the coverage performance across all algorithms is rather similar. 
This can be attributed to the ease of covering the feature space on these synthetic BBOB task where the feature dimensions are directly linked to the parameter dimensions.
This makes Random Sampling and MAP-Elites strong upper baselines on these tasks.

The Arm task poses another challenge as there is an unclear pattern connecting features and parameters, and as some regions of the feature space are unreachable.
Despite these challenges, \algoname{} competitively performs QD optimization on this task (third column Figure~\ref{fig:in-context-qd-results}).
In the Arm, the solutions with high fitness (i.e. solutions with low variance in joint angles) have a clear pattern in which similar parameter values would give high fitness.
\algoname{} very quickly identifies and extracts this pattern from the provided archive context and finds the region of high-performing solutions on this task (see third row on Figure~\ref{fig:in-context-qd-results} Max Fitness).
In terms of coverage of the feature space, \algoname{} manages to reach performance similar to MAP-Elites, showing its ability to discover the non-trivial pattern connecting parameters to features in this task. 
We also find that, as \algoname{} uses the entire archive and generates solutions based on pattern matching, its optimization dynamics and paths are very different from those of conventional QD algorithms. 
This interesting observation is most evident in this Arm task as shown in Figure~\ref{fig:archive-optimization-dynamics}.

Lastly, we demonstrate that \algoname{} is also an effective QD approach on the Hexapod (policy search) task, where both fitness and features are not in any direct way translatable from the parameter space.
This task also consists of $D=24$ parameters which is more than double the size of dimensions of previous tasks showing that \algoname{} can also be effective at higher dimensions.
Overall, \algoname{} provides an interesting perspective and promising approach for the effective generation of solutions in QD by considering an AI generator that can utilize the knowledge and context of the entire state of the archive.

\begin{figure}[t]
    \centering
    \includegraphics[width=\linewidth]{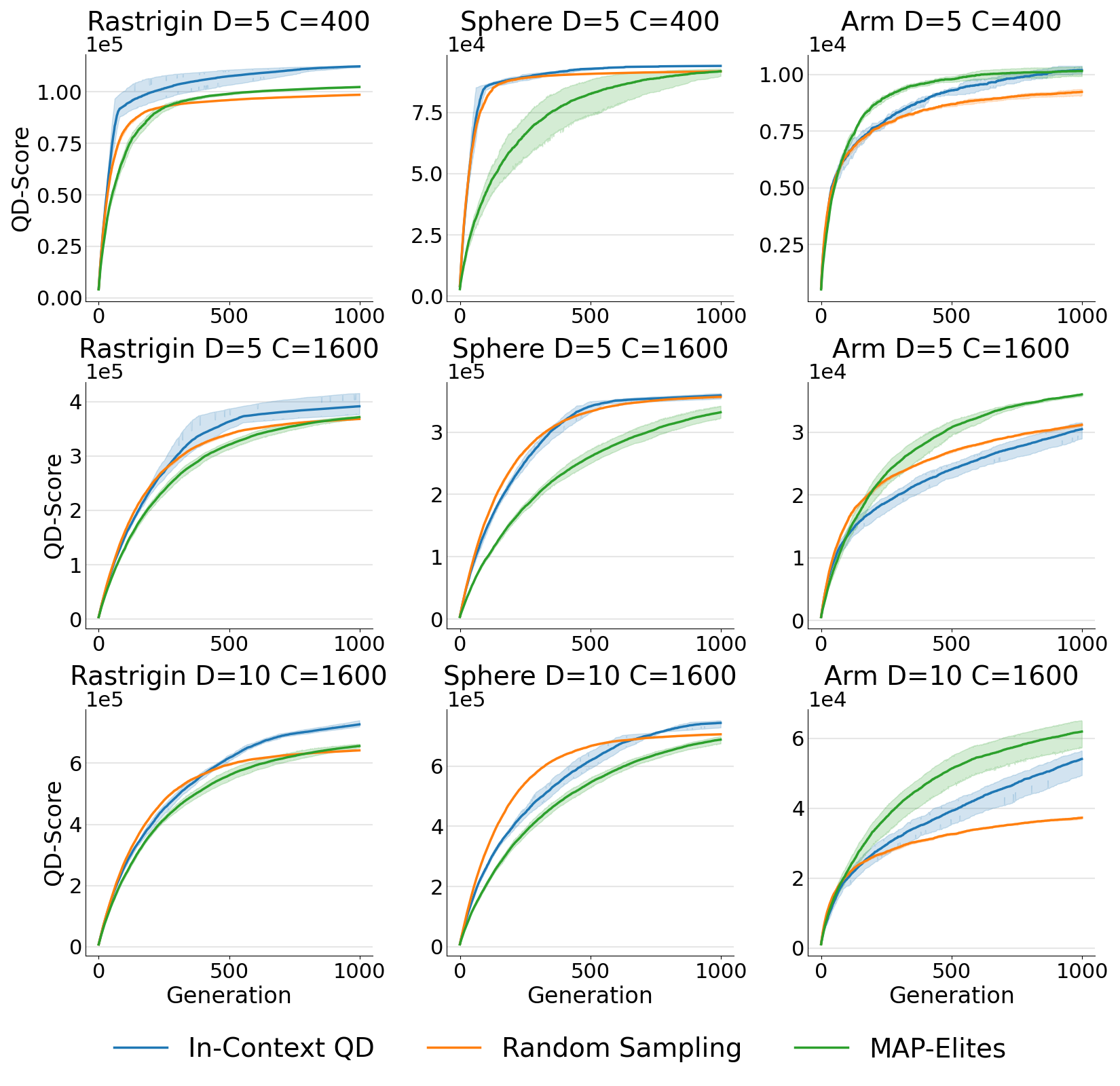}
    \captionsetup{skip=1pt}
    \caption{QD-Score of \algoname{} across a variety of parameter space dimensionality $D$ and archive sizes $C$.}
    \label{fig:main-metrics-2}
    \vspace{-2mm}
\end{figure}

In addition to the variety of different QD tasks and domains, the results presented in Figure~\ref{fig:main-metrics-2} show that \algoname{} is a general QD approach and is also robust across both archive sizes $C \in \{400, 1600\}$ and parameter search space dimensions $D \in \{5, 10\}$, performing competitively against compared baselines.



\subsection{Effect of prompt template}
We study the importance of the prompt template used for \algoname{} by comparing to existing methods and baselines for this as below:

\begin{itemize}[noitemsep]
    \item \textbf{LMX}~\citep{meyerson2023language} which prompt contains only the parameters of each solution. 
    \item \textbf{Fitness}~\citep{yang2023large, generalpatternmachines2023, lange2024large} which prompt contains the fitness score followed by the parameter. This template is generally used in conventional single-objective optimization.
    \item \textbf{Feature} which prompt consists of feature characterizations before the parameter vector. 
    \item \textbf{QD} which prompt consists of both the fitness scores and the feature as explained in the methods.
\end{itemize}

Table~\ref{tab:prompt-template} provides examples of the templates used for each of the strategies.
For all of the above methods in our study, we use the same context-building and query strategies. 
However, for LMX, the query does not affect or contribute to the prompt as it does not rely on queries.

\begin{figure}[t]
    \centering
    \includegraphics[width=\linewidth]{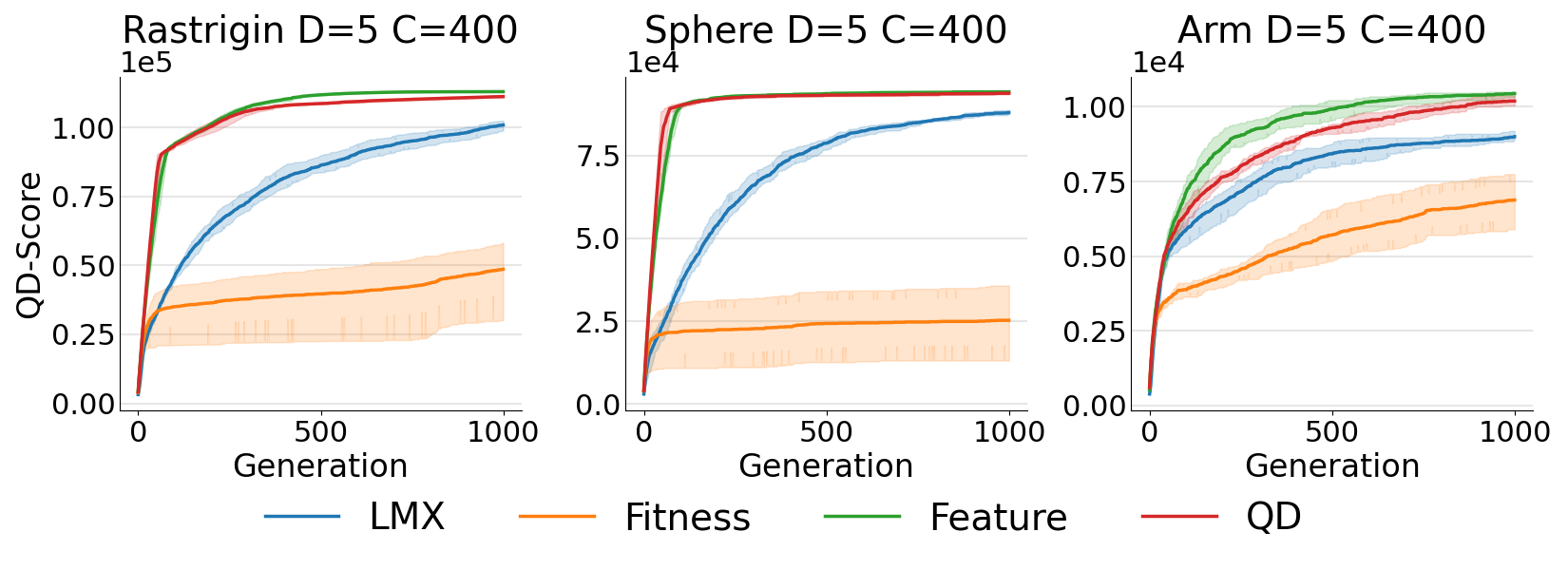}
    \captionsetup{skip=1pt}
    \caption{Performance comparison between different prompt templates (see Table~\ref{tab:prompt-template}).}
    \label{fig:context-template-study}
    \vspace{-2mm}
\end{figure}

Figure~\ref{fig:context-template-study} shows the QD-score result of this comparison where we see the importance of both the feature and descriptor in the template of the prompt.
As expected, while the fitness-only template does well at finding the best solution, it has no incentive to generate solutions with new descriptors, leading to low coverage and overall poor QD performance.
The score and label agnostic approach of LMX does better as it is not biased by any values, relying on its context to estimate a distribution from the parameter space and generating solutions from this distribution~\citep{meyerson2023language}.
In the context of QD and our experiments, since the context consists of solutions and elites from the archive, this distribution is most likely that of the hyper-volume of elites~\citep{mouret2015illuminating, vassiliades2018iso}.
Most importantly, we observe that the presence of the features $\textbf{d}$ in the template is critical to the performance of \algoname{}.
Along with the querying strategy, templates with features $\textbf{d}$ in the prompt significantly do better on QD problems.

\subsection{Effect of context structure}

\begin{figure}\centering\includegraphics[width=0.92\linewidth]{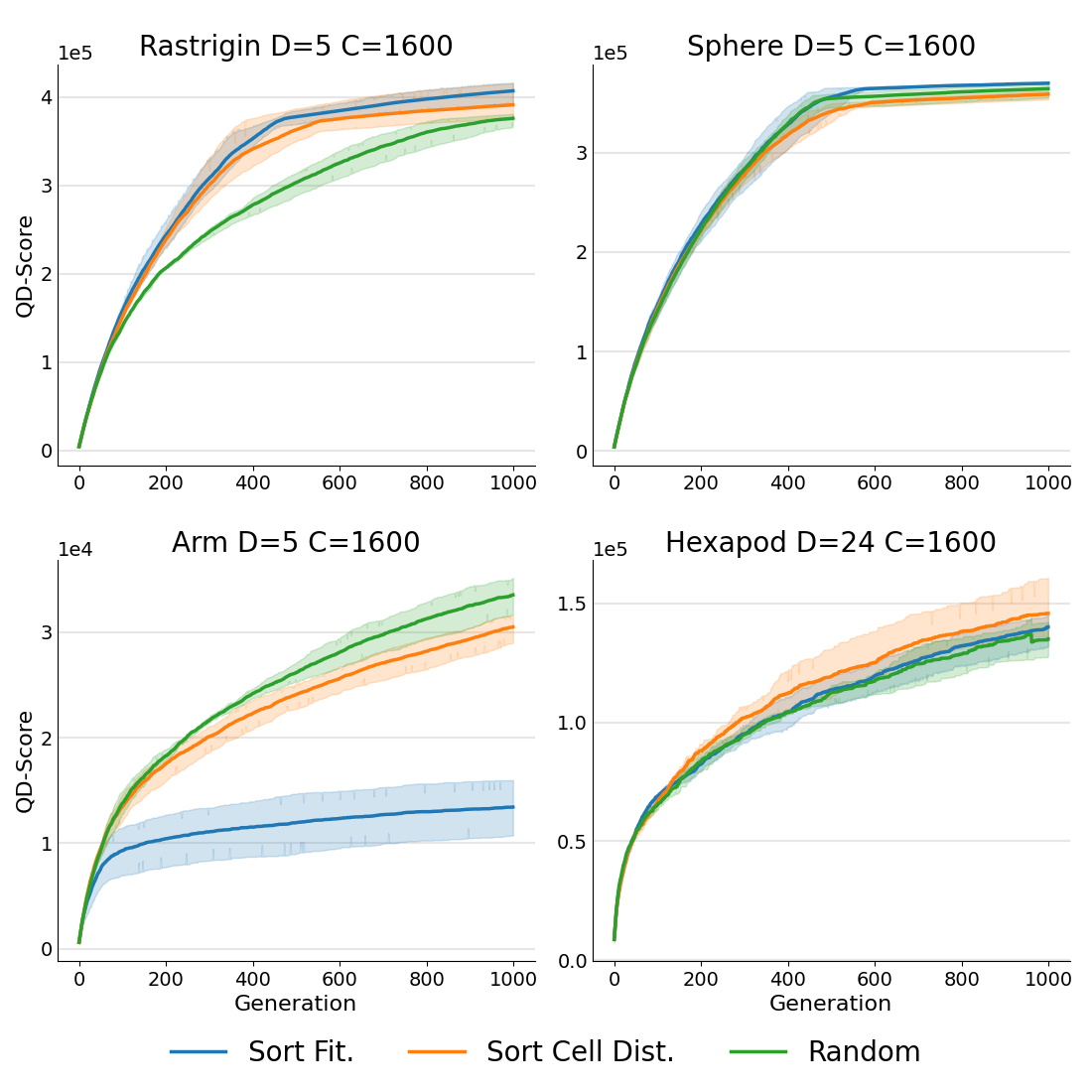}\captionsetup{skip=1pt} 
    \caption{Effect of using different methods to sort and arrange the context structure on the QD-score performance of \algoname{}.}
    \vspace{-5mm}
    \label{fig:context-structure-study}
\end{figure}

Next, we also consider how the structure of the context affects the performance of \algoname{}.
In Figure~\ref{fig:context-structure-study}, we can observe sorting according to the cell distance from the query feature performs most consistently well across different tasks.
This structure complements the feature $\textbf{d}$ in the template providing a clear trend and pattern which enables good exploration.
Interestingly, we find that the conventional sorting by fitness approach struggles in some domains especially when there is a clear pattern for high-performing solutions such as in the Arm task.
Such a structure causes the LLM to exploit the patterns of high fitness resulting in no exploration of the feature space.

\begin{table}[t!]
    \centering
    \begin{tabular}{r|l}
    \toprule
    Strategy & Template \\
    \midrule
    \addlinespace
     LMX            &  $p^i_1, ..., p^i_D$ \\
     Fitness        & $f^i: p^i_1, ..., p^i_D$ \\
     Feature   & $d^i_1, ..., d^i_G: p^i_1, ..., p^i_D$ \\
     QD             & $f^i: d^i_1, ..., d^i_G: p^i_1, ..., p^i_D$
     \\
     \bottomrule
    \end{tabular}
    \caption{Comparison of prompt templates where $f$ is the fitness score, $\textbf{d}$ are the features and $\textbf{p}$ are the parameters of the solution.}
    \label{tab:prompt-template}
    \vspace{-2mm}
\end{table}

\subsection{Effect of context size}

\begin{figure}
    \centering
    \includegraphics[width=0.92\linewidth]{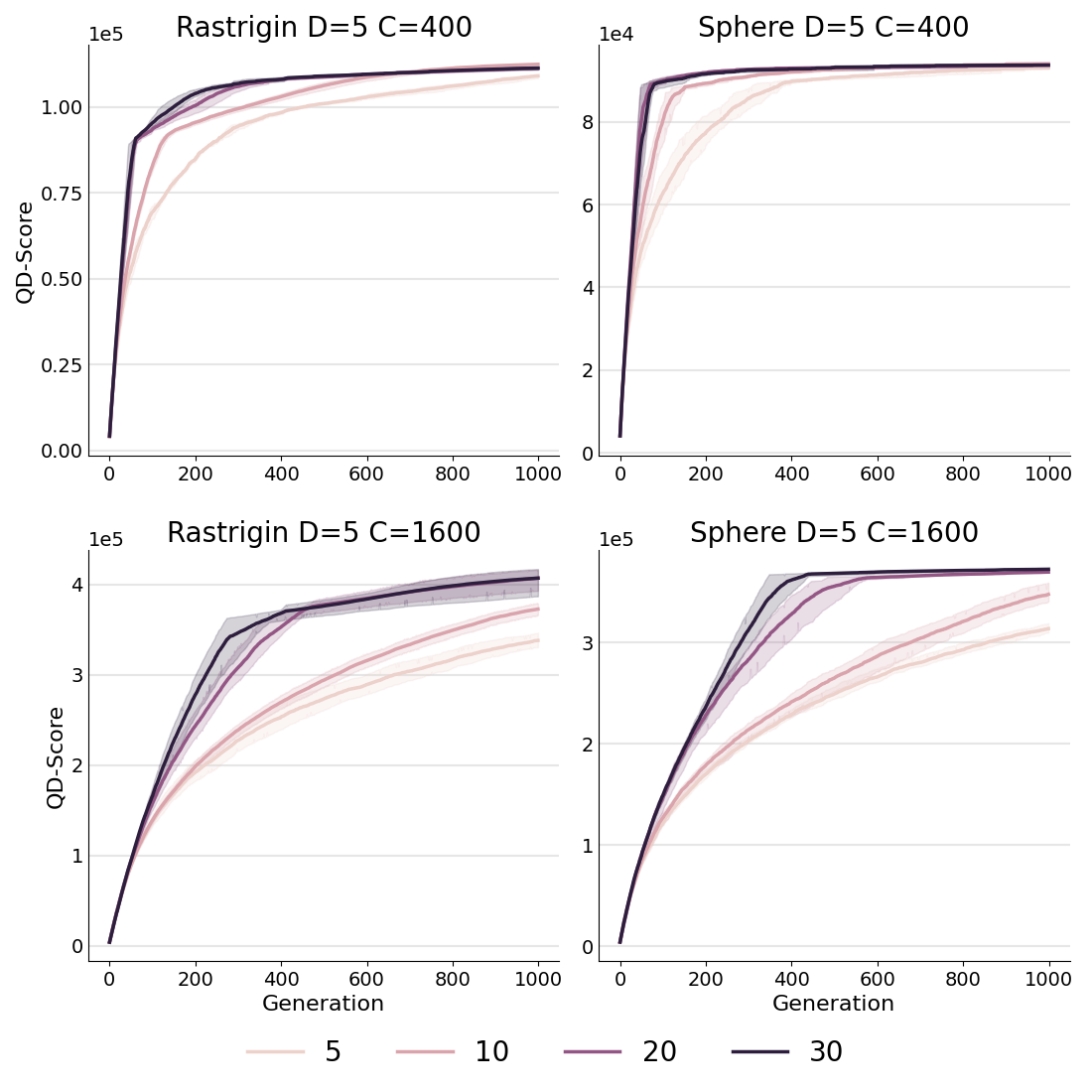}
    \captionsetup{skip=0pt}
    \caption{QD-Score performance over a range of context sizes of \algoname{}.
    We find that increasing context size helps performance.}
    \label{fig:context-size-study}
    \vspace{-3mm}
\end{figure}

A critical component of our \algoname{} is the use of the entire archive for the generation of the next solutions.
Ideally, we would like to provide AI generators with the entire diversity of the archive as context to be able to more effectively understand the distribution of the space of solutions and best generate novel and better-performing solutions w.r.t the current archive.
However, as explained in the methods, the context length of current LLMs limits the number of solutions that can fit into the context size. 
Here, we investigate the effect of the different context sizes on \algoname{} when sampling is done to approximate the entire archive of elites.
Figure~\ref{fig:context-size-study} shows that increasing the context size always helps to improve performance.
This is particularly important when the archive has a large number of niches.
We can observe in Figure~\ref{fig:context-size-study} that the difference in performance between a context size of $10$ and $20$ is more significant when $C = 1600$ compared to when $C = 400$.
At higher archive sizes, coverage is more difficult, hence getting a better distribution estimate of the feature space through more examples is beneficial.
This observation corroborates the hypothesis that sampling from the archive provides an approximation of the archive of elites, and we can get increased performance as this approximation gets better.
However, when sampling, these results indicate that there is a minimum context size that is critical to good performance and dependent on the maximum archive size $C$.

\section{Discussion and Conclusion}
In this work, we introduce \algoname{} which combines the large diversity of the QD archive to more effectively generate new solutions for QD by leveraging the in-context learning capabilities of pre-trained LLMs.
We demonstrate that this allows for strong and competitive performance on a variety of QD tasks ranging from BBO to policy search, while also showing its robustness and generality across a range of archive sizes and parameter space dimensions.
Finally, we highlight and show that the choice of the different prompt components such as the prompt template, context structure, context size, and query strategy is critical to using LLMs for QD.

In our work, we utilize LLMs for what these models thrive at and were trained to do, which are their capabilities as general pattern matching machines~\citep{brown2020language, generalpatternmachines2023} to spot patterns in the sequences and potentially complete and improve them.
Using them in this manner allows us to treat them as idea or solution generators that rely on examples that are high-quality and diverse provided through the library of solutions from QD.
Using them in this optimization loop also means that these "ideas" can be "verified" when applied and tested on the task or problem.
This is in contrast to many other applications where they are attempted to be used as sources of truth for factual answers, where they behave more like approximate retrieval machines which can at times lead to what is commonly referred to as "hallucinations".
In our setting, we rely on these "hallucinations" to dream and generate novel and better ideas or solutions given a set of examples, for which we can then test and experiment on the problem, similar to an engineer, inventor or scientist.

While we focus on the general pattern-matching abilities of LLMs for general optimization problems, they are commonly used for its abilities on mediums such as language or code that it possesses from its training data.
The strengths of these models in these domain-specific capabilities are complementary to our proposed approach and we hope to explore these applications.
This further demonstrates the versatility of pattern-matching based LLM optimization on a wide set of settings.
However, it is important to note that despite the ease and versatility of \algoname{} and its rapid exploitation of patterns, it can also struggle where patterns in the fitness, features and descriptors are unclear.
Our work highlights generating solutions using in-context learning with LLMs as just an alternative form of optimization based on noisy pattern-matching rather than gradients or random un-directed mutations that can be useful in certain settings, and not something that will dominate and eat up all QD approaches.
A more complete approach points to perhaps using \algoname{} to complement a multi-emitter setting~\citep{cully2021multi}, where the LLM generator can be an emitter which exploits patterns in problems as it arises.

\textbf{Limitations and Future Work.}
A limitation of our current approach is that we assume a defined feature space where we can query desired features. 
This is true and a controllable aspect of many QD algorithms and applications. 
However, to extend this to a fully open-ended setting where we get continuously evolving and new features, an interesting option would be to get the LLMs themselves to output the query features especially in environments and tasks that are fully open-ended that don’t have a specified feature space.
We hope to explore this in future work.

Additionally, we are also still limited to optimization problems of smaller parameter dimensions due to the context size of these models. 
Solutions to these could include models that have increasingly large context-lengths up to a million tokens~\citep{reid2024gemini} or dimension batched querying~\citep{lange2024large}.
Yet, we believe the general principles of this approach apply and extend to QD with context-based generative models as solution generators.
Potential future work also includes other query and context building strategies due to its simplicity or using LLM-based single-objective optimization~\citep{yang2023large, generalpatternmachines2023, lange2024large} with novelty objectives or curiosity objectives as in novelty search~\citep{lehman2011novelty, conti2018improving}.



\footnotesize
\bibliographystyle{apalike}
\bibliography{references} 

\end{document}